%% file: root.tex
\documentclass[letterpaper, 10 pt, conference]{ieeeconf}  %

\IEEEoverridecommandlockouts                              %

\usepackage{lipsum}

\usepackage{amsfonts,amssymb,amsmath,bm}

\usepackage{graphicx}
\usepackage{siunitx}
\usepackage{glossaries}
\usepackage[font=small]{caption}

\usepackage{xcolor}
\usepackage{ifthen}
\usepackage{flushend}
\newboolean{authnotes}
\usepackage[hidelinks]{hyperref}
\usepackage{subfigure}
\usepackage{tabularx}
\usepackage{multirow}
\usepackage{multicol}
\usepackage{makecell}
\usepackage{booktabs}
\usepackage[font=small]{caption}
\usepackage{comment}
\usepackage{cite}

\setboolean{authnotes}{true}

\ifthenelse{\boolean{authnotes}}
{
\newcommand{\todo}[1]{{{\bf\color{magenta} TODO: #1}}}

}
{
\newcommand{\todo}[1]{}
\newcommand{\nf}[1]{}
\newcommand{\db}[1]{}
\newcommand{\st}[1]{}
}

\input{misc/math_commands}
\input{misc/more_math}

\let\ACMmaketitle=\maketitle
\renewcommand{\maketitle}{\begingroup\let\footnote=\thanks \ACMmaketitle\endgroup}

\makeatletter
\newcommand*\titleheader[1]{\begingroup\gdef\@titleheader{#1}\let\footnote=\thanks\endgroup}
\AtBeginDocument{%
  \let\st@red@title\@title
  \def\@title{%
  \begin{flushleft}
    \vspace{-2.0em}
    \bgroup\normalfont\small\@titleheader\par\egroup
    \vspace{-18pt}\par\noindent\rule{\textwidth}{0.1pt}
    \end{flushleft}
    \vskip0.5em\st@red@title
        }

}

\makeatother

\title{\LARGE \bf
Placing by Touching: An empirical study on the importance\\ of tactile sensing for precise object placing}

\author{Luca Lach$^{*1, 2}$, Niklas Funk$^{*3}$, Robert Haschke$^1$, Séverin Lemaignan$^{4}$, Helge Joachim Ritter$^{1}$, \\Jan Peters$^{3,5,6,7}$, Georgia Chalvatzaki$^{3,6}$
\thanks{$*$ Authors contributed equally.}\thanks{This work was supported by the European Union's Horizon 2020 Marie Curie Actions under grant no. 813713 NeuTouch, the Horizon Europe research and innovation program under grant no. 101070600 SoftEnable, the DFG Emmy Noether Programme (CH 2676/1-1), the BMBF Project Sim4Dexterity, the BMBF Project Aristotle, and the AICO grant by the Nexplore/Hochtief Collaboration with TU Darmstadt.}\thanks{$^1$ Neuroinformatics Group, Technical Faculty, Bielefeld University
{\tt\footnotesize\{llach, rhaschke, helge\}@techfak.uni-bielefeld.de}}\thanks{$^{2}$ Institut de Robòtica i Informàtica Industrial, CSIC-UPC}\thanks{$^{3}$ Computer Science Department, Technical University Darmstadt, $^4$ PAL Robotics, $^{5}$~German Research Center for AI (DFKI), Research Department: Systems AI for Robot Learning, $^{6}$~Hessian.AI, $^{7}$~Centre for Cognitive Science.
{\tt\footnotesize\{niklas.funk, jan.peters, georgia.chalvatzaki\} @tu-darmstadt.de}}
\thanks{This work has been submitted to the IEEE for possible publication. Copyright may be transferred without notice, after which this version may no longer be accessible.}
}

\let\oldtwocolumn\twocolumn
\renewcommand\twocolumn[1][]{%
    \oldtwocolumn[{#1}{
    \begin{center}
    \vspace{-0.4cm}
        \begin{minipage}{0.245\textwidth}
            \includegraphics[width=\textwidth]{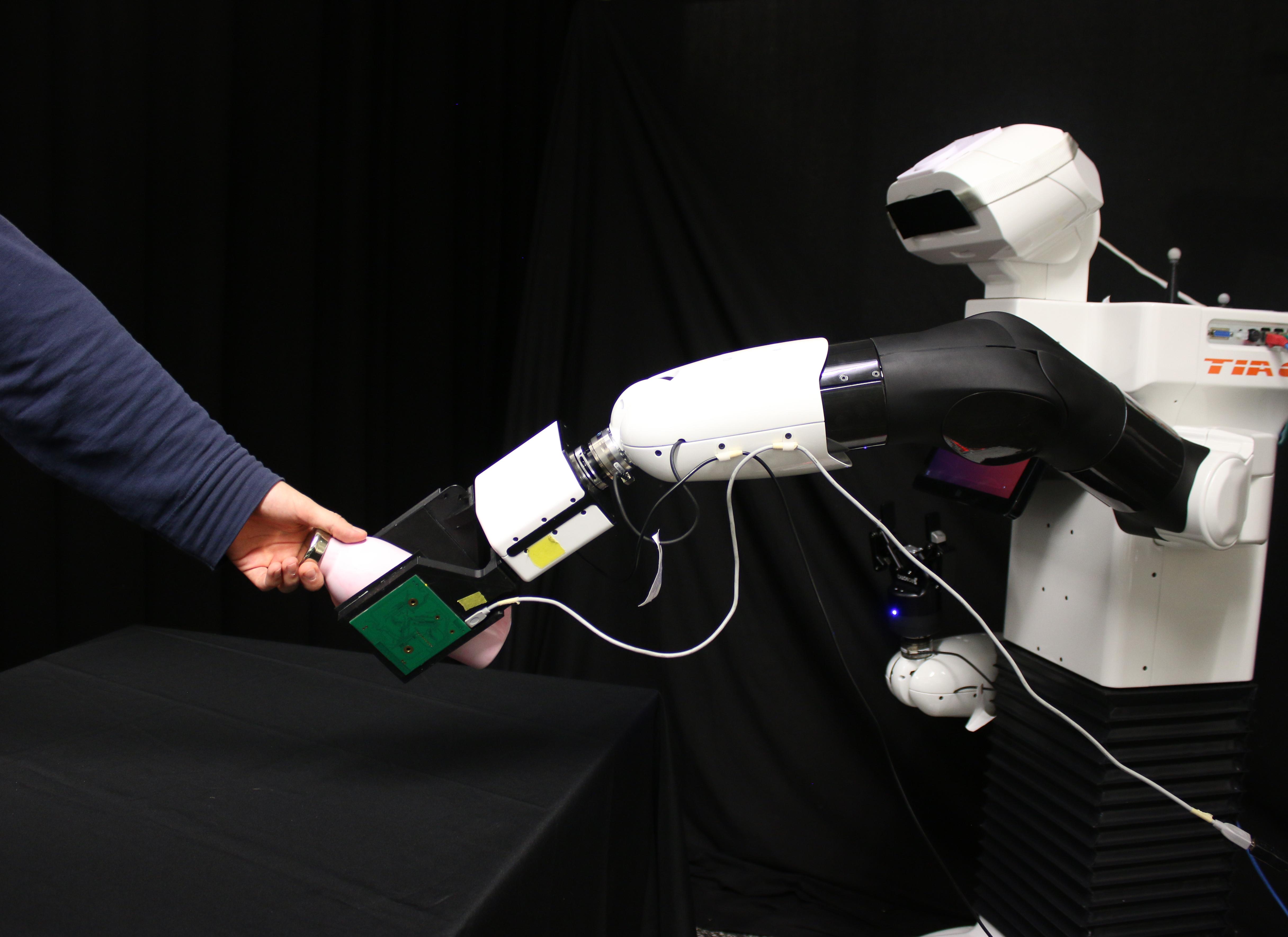}
        \end{minipage}
        \hfill
        \begin{minipage}{0.245\textwidth}
            \includegraphics[width=\textwidth]{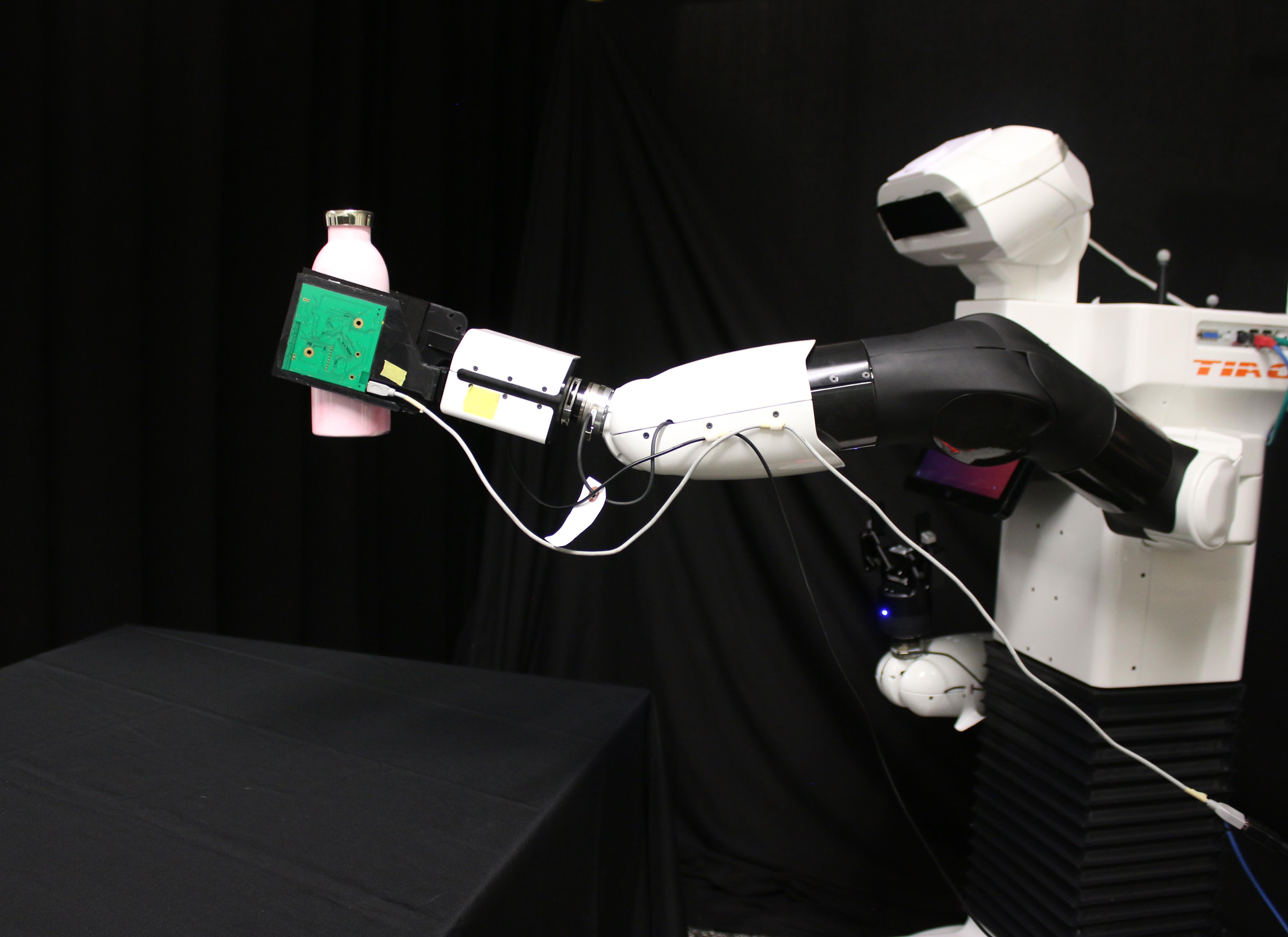}
        \end{minipage}
        \hfill
        \begin{minipage}{0.245\textwidth}
            \includegraphics[width=\textwidth]{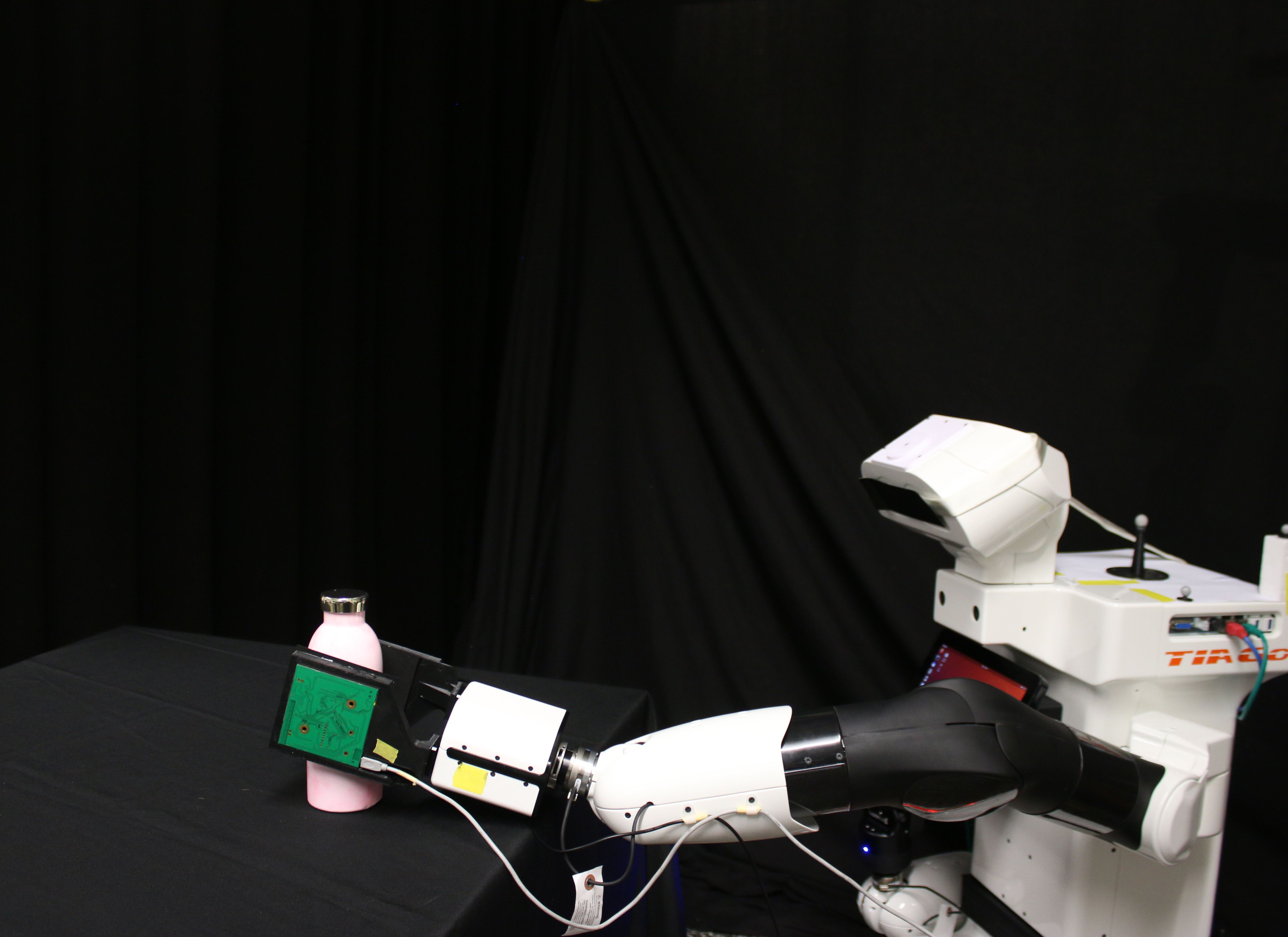}
        \end{minipage}
        \begin{minipage}{0.245\textwidth}
            \includegraphics[width=\textwidth]{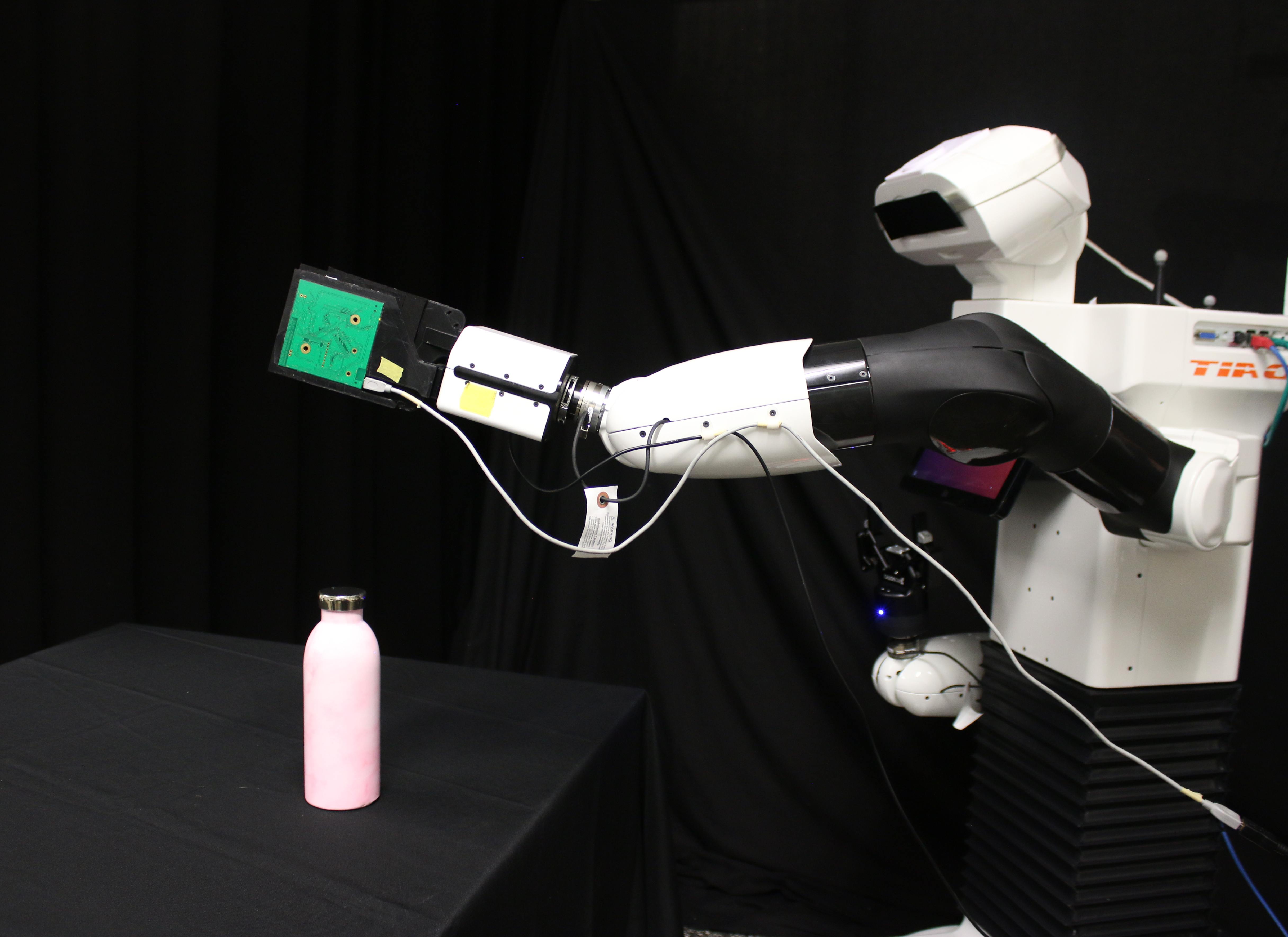}
        \end{minipage}
        \vspace{-0.1cm}
        \captionof{figure}{Stable object placing sequence with a \textit{blind} robot, from left to right: (1) an object is handed to the robot in an unknown pose. Subsequently, by leveraging the tactile sensor readings inside the gripper, we estimate how the object needs to be reoriented for stable placement. (2) Given the estimate, a controller aligns the object with the placing surface. (3) We move the robot down to place the object on the table. Contact is again detected using tactile sensors. (4) The gripper is opened, and the robot retracts.}\label{fig:main_fig}
    \end{center}
    }]
}

\titleheader{\textit{Preprint}}

\begin{document}

\maketitle
\thispagestyle{empty}
\pagestyle{empty}

\begin{abstract}
This work deals with a practical everyday problem: stable object placement on flat surfaces starting from unknown initial poses. 
Common object-placing approaches require either complete scene specifications or extrinsic sensor measurements, e.g., cameras, that occasionally suffer from occlusions. We propose a novel approach for stable object placing that combines tactile feedback and proprioceptive sensing. We devise a neural architecture that estimates a rotation matrix, resulting in a corrective gripper movement that aligns the object with the placing surface for the subsequent object manipulation.
We compare models with different sensing modalities, such as force-torque and an external motion capture system, in real-world object placing tasks with different objects. 
The experimental evaluation of our placing policies with a set of unseen everyday objects reveals significant generalization of our proposed pipeline, suggesting that tactile sensing plays a vital role in the intrinsic understanding of robotic dexterous object manipulation.
Code, models, and supplementary videos are available on \url{https://sites.google.com/view/placing-by-touching}.
\end{abstract}

\input{sections_new/intro}

\input{sections_new/related}

\input{sections_new/method}

\input{sections_new/experiments}

\input{sections_new/conclusions}

\bibliographystyle{IEEEtran}
\bibliography{references.bib}
\end{document}

%% file: misc/math_commands.tex
\def\1{\bm{1}}

\DeclareMathAlphabet{\mathsfit}{\encodingdefault}{\sfdefault}{m}{sl}
\SetMathAlphabet{\mathsfit}{bold}{\encodingdefault}{\sfdefault}{bx}{n}

\makeatletter
\newcommand{\StatexIndent}[1][3]{%
  \setlength\@tempdima{\algorithmicindent}%
  \Statex\hskip\dimexpr#1\@tempdima\relax}
\makeatother

%% file: misc/more_math.tex
\newcommand{\zp}{\vec{z}_p}
\newcommand{\zpg}{\vec{z}^{G}_p}
\newcommand{\zpgp}{\vec{z}^{G'}_p}
\newcommand{\zgt}{\vec{z}^{gt}_p}
\newcommand{\zs}{\vec{z}_s}
\newcommand{\zsg}{\vec{z}^G_s}
\newcommand{\zsgp}{\vec{z}^{G'}_s}

\newcommand{\Rot}{\mathbf{R}}
\newcommand{\Rwo}{\Rot^W_O}
\newcommand{\Rwop}{\Rot^W_{O'}}

\newcommand{\Rggp}{\Rot^G_{G'}}

\newcommand{\Rwg}{\Rot^W_G}

\newcommand{\Rgw}{\Rot^G_W}
\newcommand{\Roop}{\Rot^O_{O'}}
\newcommand{\Rgop}{\Rot^G_{O'}}

%% file: sections_new/intro.tex
\section{Introduction}
Human dexterity is impeccable and is largely attributed to the human sense of touch. Tactile sensing enables precise, reliable, dexterous manipulation, a crucial component for the vision of generalized autonomy and of more capable, intelligent autonomous robotic systems \cite{prassler2012towards}.
Integrating the sense of touch is a key research topic in robotics, with prominent works including tactile insertion \cite{dong2019tactile, dong2021tactile, kim2022active}, in-hand manipulation \cite{yousef2011tactile, van2015learning, ward2017model, funk2021benchmarking}, assembly \cite{belousov2022robotic, funk2022learn2assemble, funk2022graph}, human-robot-interaction \cite{bhattacharjee2013tactile, dean2017tomm}, and object pose estimation \cite{alvarez2019visual, villalonga21a}. 

In the past few years, significant effort has been put into designing sensitive and compact tactile sensors that allow easy integration with robotic systems.
Compared to more traditional Force/Torque (F/T) sensors that only provide one 6-dimensional F/T measurement at the sensor's location, tactile sensors offer a high spatial resolution and measurements directly at the points of contact.
However, the high-dimensional tactile signals are usually unsuitable for direct integration in control loops and require additional preprocessing.
Tactile sensors can be realized using a wide range of sensing principles \cite{KAPPASSOV2015195}.
Recently, tactile sensors based on piezo-resistive sensing distributed in an array of taxels \cite{piezoresistive1988, myrmexpaper, tactiledataglove2012, vidal2011three, kerpa2003development}, and those relying on cameras capturing a soft gel's surface  \cite{visionbasedtactile2004, ward2018tactip, lambeta2020digit, donlon2018gelslim, sferrazza2019design, romero2020soft} have become increasingly popular amongst the robotics community.

Herein, we focus on piezo-resistive tactile sensors. In particular, we use a pair of Myrmex tactile sensors \cite{myrmexpaper}, which feature a $16 {\times} 16$ taxel array sampled at 1 kHz, i.e., matching the frequency of good F/T sensors, whilst visuotactile sensors can perform at best around 90 Hz. Regarding costs, tactile sensors like Myrmex or GelSight are considerably cheaper than F/T sensors.

This work studies the benefit of local tactile measurements between gripper and object for stable and reliable object placing.
Stable object placement is an essential skill for any autonomous robotic system, particularly for capable assistive household robots. It forms the basis for many tasks, such as object rearrangement, assembly, sorting, and storing goods.
While a large body of prior works exists on stable object placing \cite{jiang2012learning, harada2012object, ma2018regrasp, mitash2020task, manuelli2019kpam, haustein2019object, newbury2021learning}, none of those works investigates the contribution of tactile feedback in stable placing. Rather, they rely either on vision systems, which are prone to occlusions and require external sensors, or accurate scene descriptions, which demand cumbersome manual labor.
We attempt to fill this gap by investigating the impact of tactile sensing in this simple yet challenging scenario.

We propose an effective pipeline for translating taxel-based measurements into useful features for learning a pose correction signal to ensure optimal object placement.
A placing action is optimal if the object's placing normal (orthogonal to its placing surface) is colinear with the normal of the placing surface, e.g., a table.
Our method comprises a deep convolutional neural network that predicts a corrective rotation action for the gripper. Given the current tactile sensor readings and potentially adding other signals, e.g., F/T information, we predict a rotation matrix w.r.t. the current gripper frame (cf. Fig. \ref{fig:frames}). The z-axis of this predicted frame corresponds to the object's placing normal. This prediction is subsequently used to plan a hand movement to align the object's placing normal with the table's normal.
After this single-step prediction and alignment, we attempt to place the object on the surface while keeping the previously determined orientation fixed.
The major challenge here is to predict the object's placing normal solely from tactile and proprioceptive sensors instead of employing traditional extrinsic vision-based methods.
To assess the importance of learning-based placing policies for this problem, we compare our method to two classical baseline approaches.

Our main contributions are twofold; (i) the development and training of tactile-based policies for stable object placing without requiring any extrinsic visual feedback, and (ii) an open-source suite of our dataset, CADs, pretrained models, and the codebase of all methods (both classical and deep learning ones) from our extensive real-robot experiments. 

Overall, our study confirms that tactile sensing can be a powerful and valuable low-cost addition to robotic manipulators: their signals provide features that increase reliability and robot dexterity.

%% file: sections_new/related.tex
\section{Related work}

\noindent\textbf{Object placing.} Stable object placing is a crucial skill for autonomous robotic systems.
Many prominent tasks in the robotic community, such as object rearrangement or assembly, require robotic pick \& place sequences that heavily rely on this skill.
The authors of \cite{harada2012object} propose a model-based pointcloud-conditioned approach for stable object placing by matching polygon models of object and environment.
Similarly, \cite{jiang2012learning} uses pointcloud observations for extracting meaningful feature representations for learning to place new objects in stable configurations.
More recently, \cite{manuelli2019kpam} proposes to exploit learned keypoint representations from raw RGB-D images for solving category-level manipulation tasks. \cite{mitash2020task} also uses a combination of vision and learning for manipulating unknown objects.
\cite{haustein2019object} presents a planning algorithm for stable object placement in cluttered scenes requiring a fully specified environment. Closely related to our work is \cite{newbury2021learning}, which presents an iterative learning-based approach for placing household objects onto flat surfaces but using a system of three external depth cameras for input.

While most of these works deal with the problem of generating stable placing poses for unknown objects, the main difference to our work lies in the input observation--none of them considers tactile sensing. Instead, they all rely on single or even multiple depth/RGB images. Relying on image data might be problematic due to gripper-object occlusions in highly cluttered scenes, especially if the object ends up inside the gripper without any prior knowledge of its pose. 
Additionally, external sensing systems require careful and precise calibration w.r.t. the robot, which is often tedious, time-consuming, and error-prone.
In contrast, tactile sensors directly provide the contact information between the object and gripper, independent of the surrounding environment.

\noindent\textbf{In-hand object pose estimation.}~Due to the inherent difficulties of estimating a grasped object's pose and due to its importance for tasks like pick \& place or in-hand manipulation, multiple methods for object-in-hand pose estimation have been developed.
The authors of \cite{bimbo2016hand} solely exploit tactile sensors and match their signal with a local patch of the object's geometry, thereby estimating its pose.
Other works \cite{alvarez2017tactile, pfanne2018fusing} make use of both visual and tactile inputs. While \cite{alvarez2017tactile} only requires an initial visual input for initializing a particle filter, \cite{pfanne2018fusing} employs an extended Kalman filter constantly using vision \& touch.
Recent progress in deep learning has fostered data-driven methods for in-hand object pose estimation. \cite{liu2021semi, doosti2020hope, sodhi2022leo} present end-to-end approaches based on RGB images. While \cite{liu2021semi, doosti2020hope} directly outputs pose predictions, \cite{sodhi2022leo} learns observation models that can later be exploited in optimization frameworks. \cite{anzai2020deep} fuses vision and tactile in an approach that self-determines the reliability of each modality, while \cite{kelestemur2022tactile} exploits a learned tactile observation model in combination with a Bayes filter.
Following the successes of recent deep learning approaches, we propose to learn an end-to-end direct mapping from tactile input for estimating the grasped object's placing normal. We want to point out that we are not interested in estimating the object's full 6D pose. Instead, we only focus on aligning the object with the placing surface.
Moreover, our proposed method does not require any repetitive measurements or filtering and solely needs to be queried once.
Finally, our method requires a very small training set.

\noindent\textbf{Insertion.} Stable object placing is also related to tactile insertion. Successful completion of both tasks requires suitable alignment between object \& table, or peg \& hole.
Several works approach challenging insertion tasks using tactile sensors \cite{li2014localization, dong2019tactile, dong2021tactile, pirozzi2018tactile, kim2022active}.
The authors of \cite{li2014localization} leverage vision-based tactile sensors for precisely localizing small objects inside the gripper. This information is subsequently exploited for small-part insertions using classical control.
\cite{dong2019tactile, dong2021tactile} also focus on solving tight insertion tasks using learned tactile representations. Both exploit the tactile measurements as a feedback signal to predict residual control commands.
Recently, \cite{kim2022active} demonstrated tactile insertion through active extrinsic sensing, i.e., explicitly estimating the contact between a peg and a hole. 
Contrarily to tactile placing, tactile insertion requires environment interactions -- typical strategies involve highly compliant policies for guiding the insertion~\cite{beltran2020variable}.
In tactile-based object placing instead, it is crucial to estimate the properties of the currently held object for identifying stable placing poses. To the best of our knowledge, we are the first to investigate tactile sensing for the stable placing of objects onto flat surfaces, providing a thorough analysis of the different combinations of sensing modalities and comparisons with classical and deep approaches for concluding on the importance of tactile sensing when planning object-placing actions.

%% file: sections_new/method.tex
\section{Stable object placing}
This work considers one of the most common household manipulation tasks: placing on planar surfaces like shelves or tables.
For an object to be placed optimally, one of its (potentially many) placing faces need to be aligned with the placing surface prior to releasing from the hand.
Which faces of an object are safe for placement is both object- and context-dependent.
In the following, we will first describe the different phases of stable object placing, then provide a mathematical definition of the pose correction action necessary for aligning the object with the placing surface. Finally, we describe the controllers that execute the pose correction and object placing.

\begin{figure}[t]
    \centering
    \includegraphics[width=0.45\textwidth]{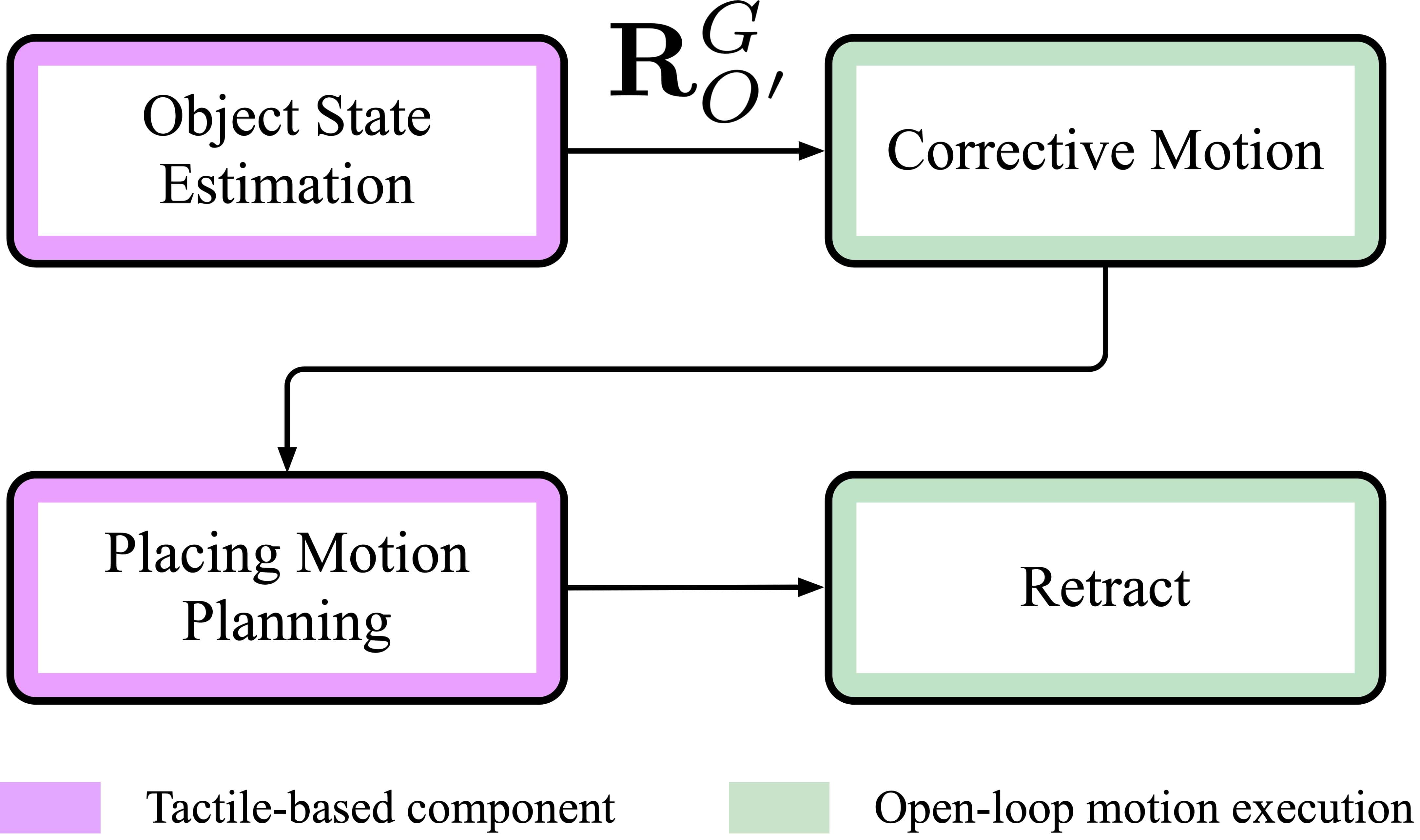}
    \caption{\small Schematic overview of the four phases of stable object placing corresponding to the sequence shown in Fig.~\ref{fig:main_fig}. Components relying on tactile data are highlighted, as well as those utilizing motion planning and execution.}
    \label{fig:flowchart}
    \vspace{-0.6cm}
\end{figure}

\subsection{The four phases of object-placing by touching}
We structure stable object placing into four phases, as shown in Figs.~\ref{fig:flowchart} \&~\ref{fig:main_fig}.
At the beginning of the object-placing pipeline, we only assume the object to be inside the gripper. 
Yet, we do not have any prior information about its pose, as would be the case after a handover.
In the first phase, given sensory information, we estimate the object's placing normal, which allows quantifying the misalignment with the placing surface (see Sec. III-B).
Based on the misalignment prediction, we compute and execute a corrective motion (see Sec. III-C).
For the third phase, a placing motion is planned and executed that moves the object towards the placing surface linearly while maintaining the object's orientation.
Once the robot detects table-object contact in the tactile responses, it transitions to the fourth phase, where it opens up the gripper and retracts.

\subsection{Definition of pose correction action for stable placing}

\begin{figure}[t]
\hfill
\subfigure[Frames and placing normal]{
    \includegraphics[width=0.46\columnwidth]{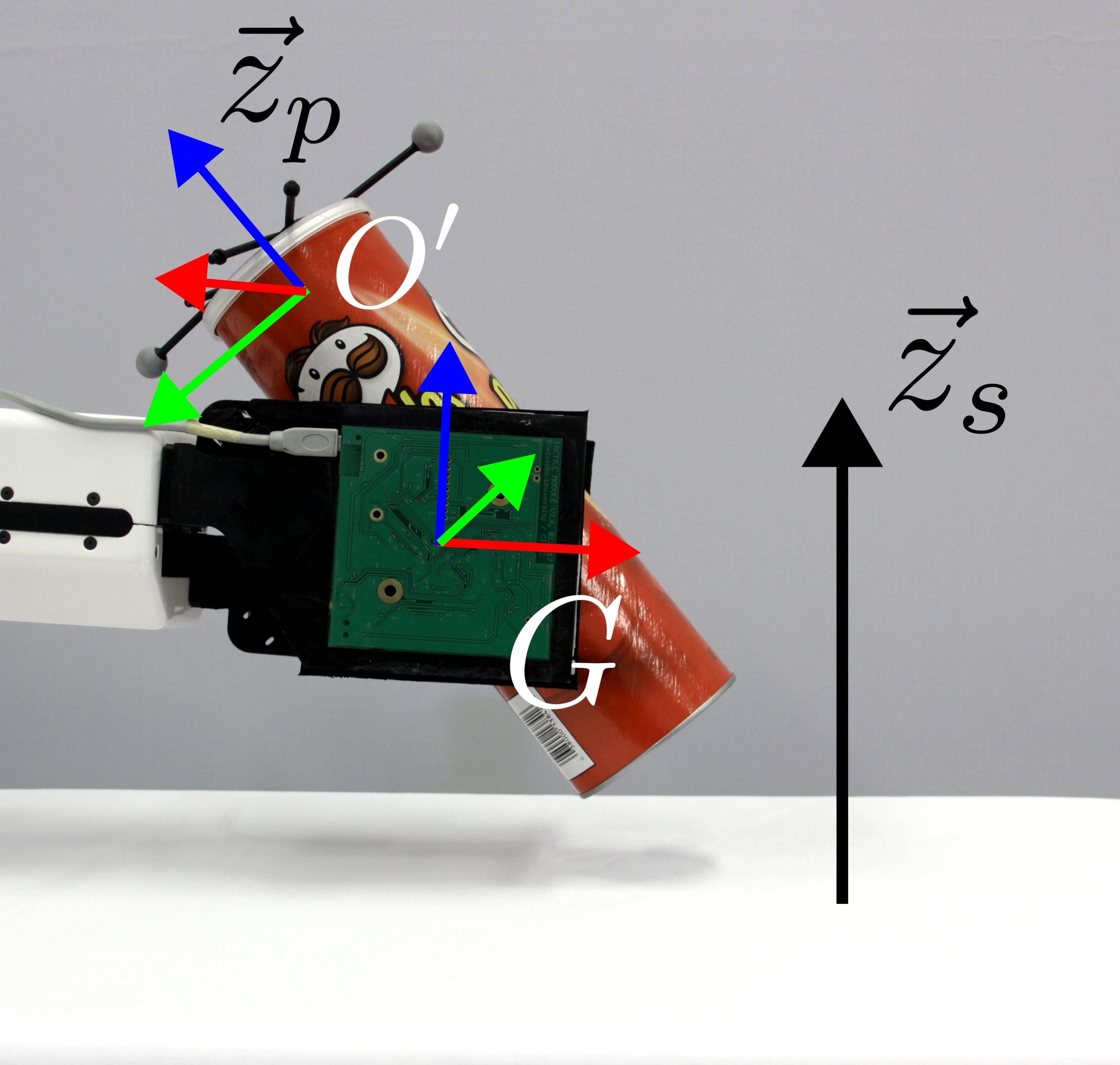}
    \label{fig:frames}
}
\hfill
\subfigure[Corrective object motion]{
    \includegraphics[width=0.44\columnwidth]{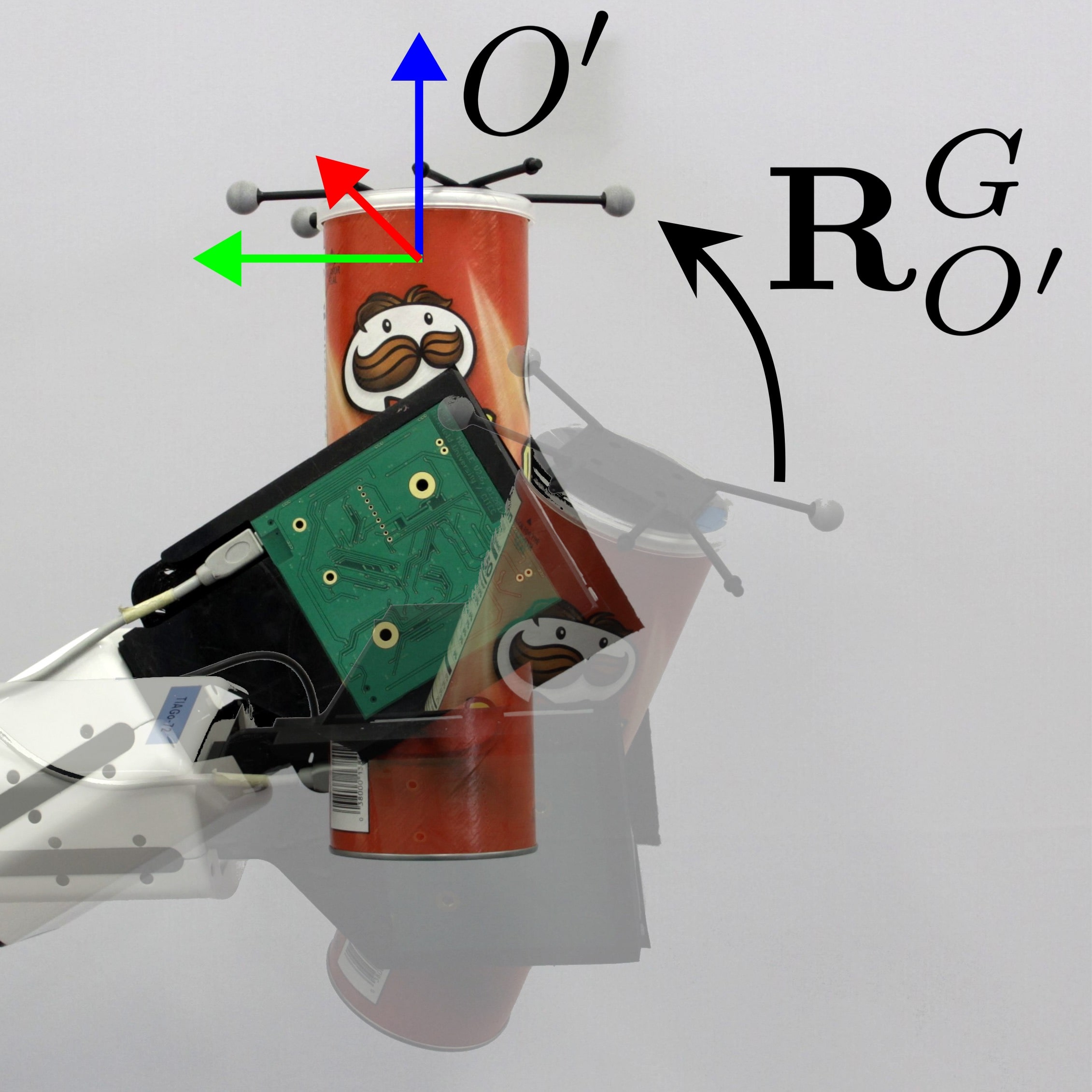}
    \label{fig:corr_mov}
}
\hfill
\caption{Illustration of the problem setting. (a) shows the gripper frame $G$, a possible object placing frame $O'$ and the surface's placing normal $\zs$. (b) Result of the corrective motion is to align the object with the placing surface based on the estimation of $\Rgop$.}
\label{fig:myrmex_data}
\vspace{-0.6cm}
\end{figure}

As illustrated in Fig. \ref{fig:frames}, we define the object's placing normal $\zp = \Rwo \Roop \vec z\,= \Rwop \vec z$ to lie along the $\vec z$-axis in the so-called local object placing frame $O'$ with the rotation matrix from a reference frame, e.g. ``world'', to the object placing frame given by $\Rwop\, \in SO(3)$. 
$\Rwo$ \& $\Roop$, thus, describe the rotation matrices from the world to the object frame, and from the object frame to the object's placing frame, respectively.
Note that we deliberately introduce this local object placing frame $O'$, in addition to the object's pose frame $O$, for two reasons. On the one hand, there might exist multiple placing frames per object, and on the other hand, to highlight that knowing the object's pose might not be informative enough for stable placing, for instance, in the case of lacking precise information about the object's geometry.
Since we only consider scenarios where the object was already grasped, $\Rwop$ can be decomposed into two distinct rotation matrices
\begin{align}
    \Rwop = \Rwg \, \Rgop,
\end{align}
where $\Rwg, \Rgop \in SO(3)$ describes the orientation of the gripper w.r.t. the world and the object's placing frame within the gripper, respectively.
While the former can be reliably estimated via forward kinematics from proprioceptive feedback, the latter is usually unknown. 
Here, we make the assumption, that one suitable placing face of the object is oriented toward the ground, which is generally the case after grasping or human-robot object handovers of various household objects.
Thus, $\Rgop$ has to be estimated based on sensor measurements as it is an indispensable ingredient for generating a motion that corrects the object misalignment and enables appropriate placing.

\subsection{Object reorientation and placing motion}

Next, we require a placing controller that, given $\Rgop$, generates two movements: a corrective arm movement that aligns $\zp$ with $\zs$ (cf. Fig. \ref{fig:corr_mov}) and a downward placing motion that stops on table contact and releases the object afterward.
For the corrective movement, we first project $\zs$ into the local gripper frame:
\begin{align}
    \zsg = \Rgw \vec z
\end{align}
Then, we calculate the rotation $\Rggp$ that rotates the current gripper frame such that in the resulting  one ($G'$), $\zpgp$ aligns with $\zsgp$. This is achieved by finding the rotation axis with $\vec{a} = \zpg \times \zsg$ and the rotation angle $\theta = \text{cos}^{-1} (\zpg \cdot \zsg)$, with the vector cross-product $\times$.
Using the rotation $\Rggp$, we can try to find an inverse kinematics (IK) solution that realizes the object reorientation.

\begin{figure}[t]
\hfill
 \subfigure[Left Myrmex Sensor]{\includegraphics[width=0.45\columnwidth]{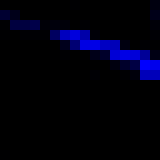}}
\hfill
\subfigure[Right Myrmex Sensor]{\includegraphics[width=0.45\columnwidth]{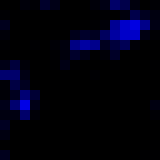}}
\hfill
\vspace{-0.2cm}
\caption{Sensor response of both Myrmex sensors while holding the cylindrical object. Each reading of the 16$\times$16 taxels corresponds to the currently measured normal force. The images were scaled up, and the force readings are amplified for visualization purposes.}
\label{fig:myrmex_data}
\vspace{-0.7cm}
\end{figure}
An IK solution that leads to a secure placing configuration should satisfy more constraints than the object reorientation alone.
After reorientation, we will execute a linear downward movement in Cartesian space to acquire table-object contact, and we need to ensure that the object will be the first point of contact with the table.
Arm configurations suitable for placing thus need to ensure that the object frame (located in the gripper) has a lower $z$-coordinate in the world frame than the wrist. %
The commonly used hierarchy-of-tasks approach \cite{slotine1991general} allows us to formulate such additional constraints for the IK.
Since many IK solvers are sensitive to the initial arm configuration, we search for solutions starting from 20 different initial poses, which are sensible placing configurations and chose the one with the lowest error that is most similar to our current arm pose.
The motion to reach this solution is generated by linear interpolation in joint space starting from the robot's current arm position.

For the placing motion, a trajectory is generated that moves the gripper linearly downward.
This is realized using TIAGo's torso joint.
During this motion, tactile measurements are continuously monitored.
We interpret a spike in tactile sensation as making contact with the table.
The torso controller is then signaled to halt execution.
Finally, the object is released by opening the gripper and moving the torso upward again.

\section{Object pose correction estimation with tactile sensing}

As motivated previously, the key component for stable object placing is the estimation of the object placing frame w.r.t. the gripper frame ($\Rgop$).
According to this transformation, the object is re-oriented prior to placing.
However, determining this quantity is difficult, as the object is handed over in an unknown pose, it is occluded by the gripper, and herein we also do not assume any prior knowledge about the object type. 
We, therefore, propose estimating $\Rgop$ from the signals of the tactile sensors inside the gripper.
The Myrmex tactile data is represented as a $16\times16$ matrix of normal force readings that are normalized in $[0,1]$.
Fig. \ref{fig:myrmex_data} shows a visualization of the tactile readings from the right and left sensors while holding an object.
Next, we, first, introduce our proposed Neural Network, and, subsequently, explain line-fitting baselines for recovering $\Rgop$ from the tactile readings.

\begin{figure}[t!]
    \centering
    \includegraphics[width=0.95\linewidth]{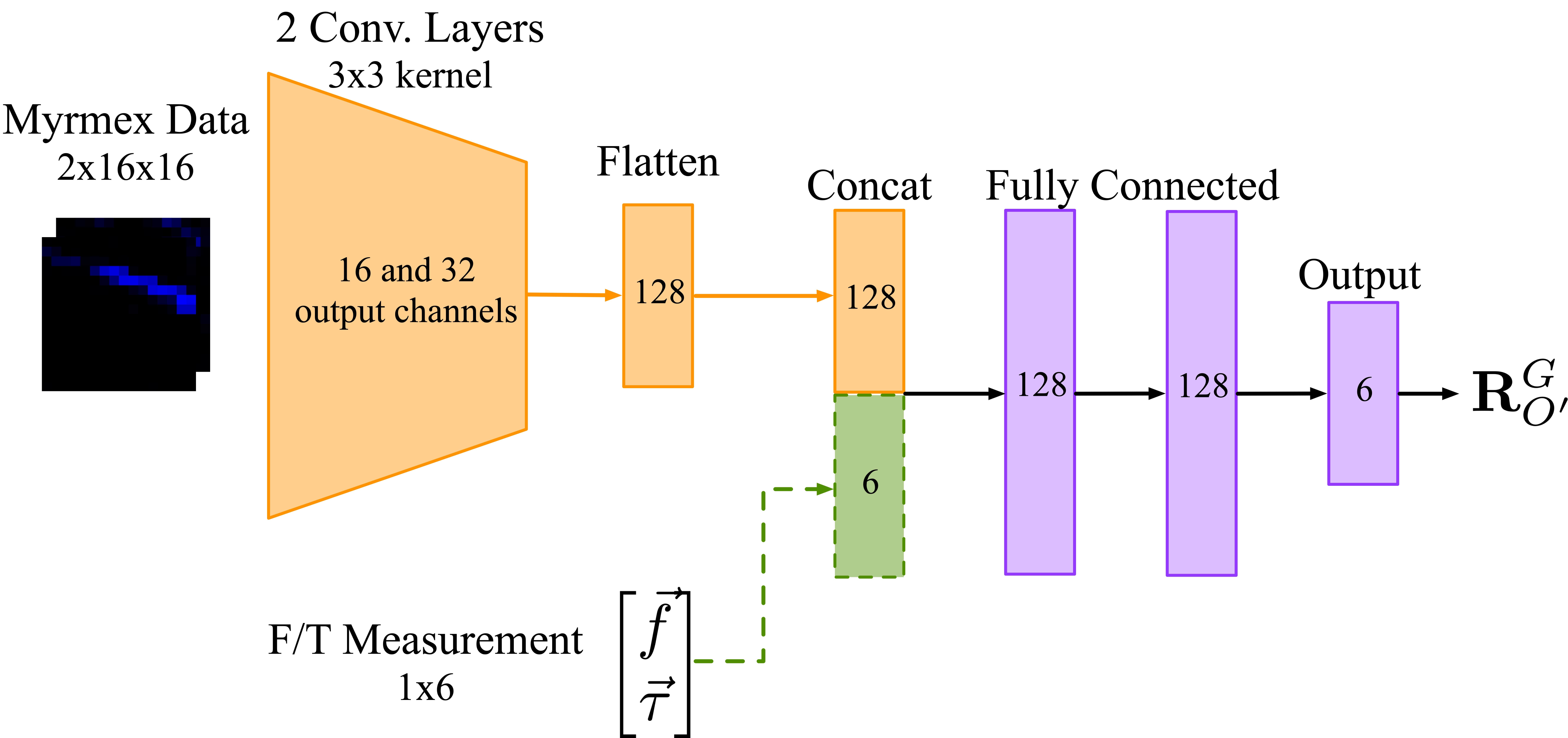}
    \vspace{-0.15cm}
    \caption{Overview of our general neural network architecture used in all models. Tactile measurements from both sides of the gripper are stacked along the channel dimension of the convolutional layers. Depending on the model type, either the tactile feature embedding, the raw F/T data, or both concatenated are fed to the MLP. Finally, the network estimates the corrective rotation $\Rgop$.}
    \label{fig:network}
    \vspace{-0.7cm}
\end{figure}

Since we require a solution that is flexible enough to deal with different objects, that can handle the sensors' noise, and convert the high-dimensional readings into a signal suitable for reorienting the objects, we propose to employ a neural network. 
The general network architecture is visualized in Fig. \ref{fig:network}.
To process the tactile data, we first use two convolutional layers with a $3 \times 3$ kernel each, and 16 and 32 output channels respectively.
The output of the last convolutional layer is then fed to a Multilayer Perceptron (MLP) consisting of two hidden layers with 128 neurons each and ReLU activation functions, followed by a dropout layer with a dropout probability of $p=0.2$.
F/T data can be optionally fed into the MLP as an additional input signal, which is concatenated with the tactile features.

To smoothly represent the rotation matrix in the output layer, we use the 6D representation comprising the first two columns of $\Rgop$, as introduced in \cite{zhou2019continuity} and has been shown to exhibit superior properties for learning in $SO(3)$.
Each estimate is then converted into an $SO(3)$ rotation matrix for the computation of the loss, which is defined as
\begin{align}
\label{eq:loss}
    \mathcal{L}(\mathbf{R}, \zgt) = \text{cos}^{-1} \Bigl( \Rwg\,\mathbf{R}\, \vec z \cdot \zgt \Bigr),
\end{align}
where $\mathbf{R} = \Rgop$ is the quantity of interest and the prediction of the network, and $\zgt$ is the ground truth measurement of the object's placing normal in the world frame that is obtained through an OptiTrack motion capture system.
By taking $\text{cos}^{-1}$, the loss lies in the interval $[0,\pi]$, and can be interpreted as the angular distance error between predicting $\zp$ using the network's output $\mathbf{R}$ and the ground truth.

%% file: sections_new/experiments.tex
\section{Experimental evaluation}

For the experimental evaluation of all models, we use the TIAGo robot from PAL Robotics, equipped with a parallel-jaw gripper.
To obtain tactile data, we exchange TIAGo's non-sensorized fingers with Myrmex sensors that are fixed to the gripper via a 3D-printed adapter.
This piezo-resistive sensor produces $16\times16$ normal force measurements with a readout frequency of 1kHz.
Each sample is represented as a 2D matrix of normalized force measurements, where we transform the raw sensor readings from $[0, 4095]$ to lie in $[0, 1]$, where values of 1 refer to the maximum force that can be measured in a sensor cell (cf. Fig. \ref{fig:myrmex_data}).

\subsection{Data collection and training}

\begin{figure}
\hfill
\subfigure[Objects used during data collection (left) and for out-of-distribution evaluation (right).\label{fig:objects}]{\includegraphics[width=0.62\columnwidth]{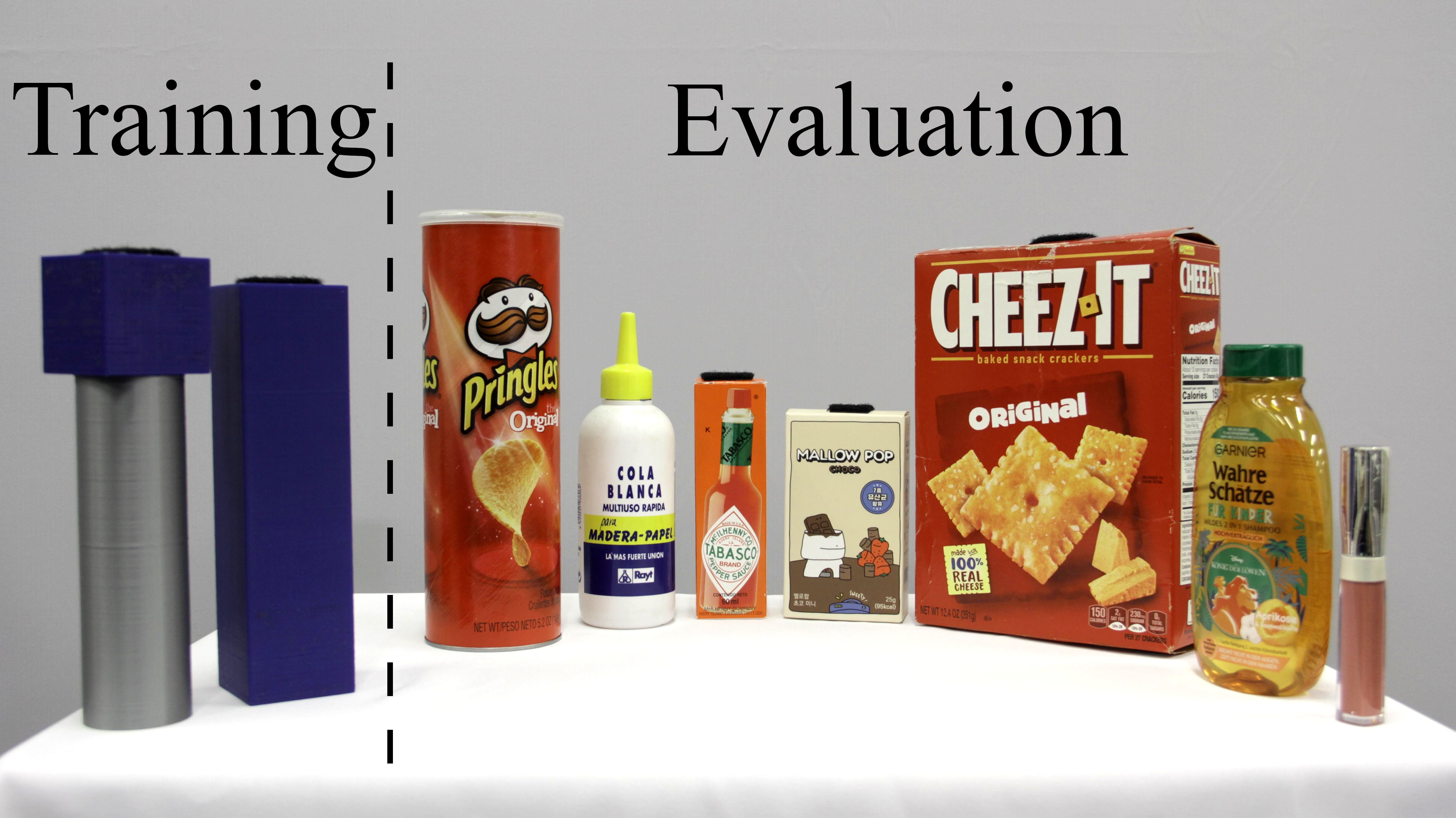} }
\hfill
\subfigure[Variations of in-hand object poses used during evaluation.\label{fig:evalposes}]{\includegraphics[width=0.35\columnwidth]{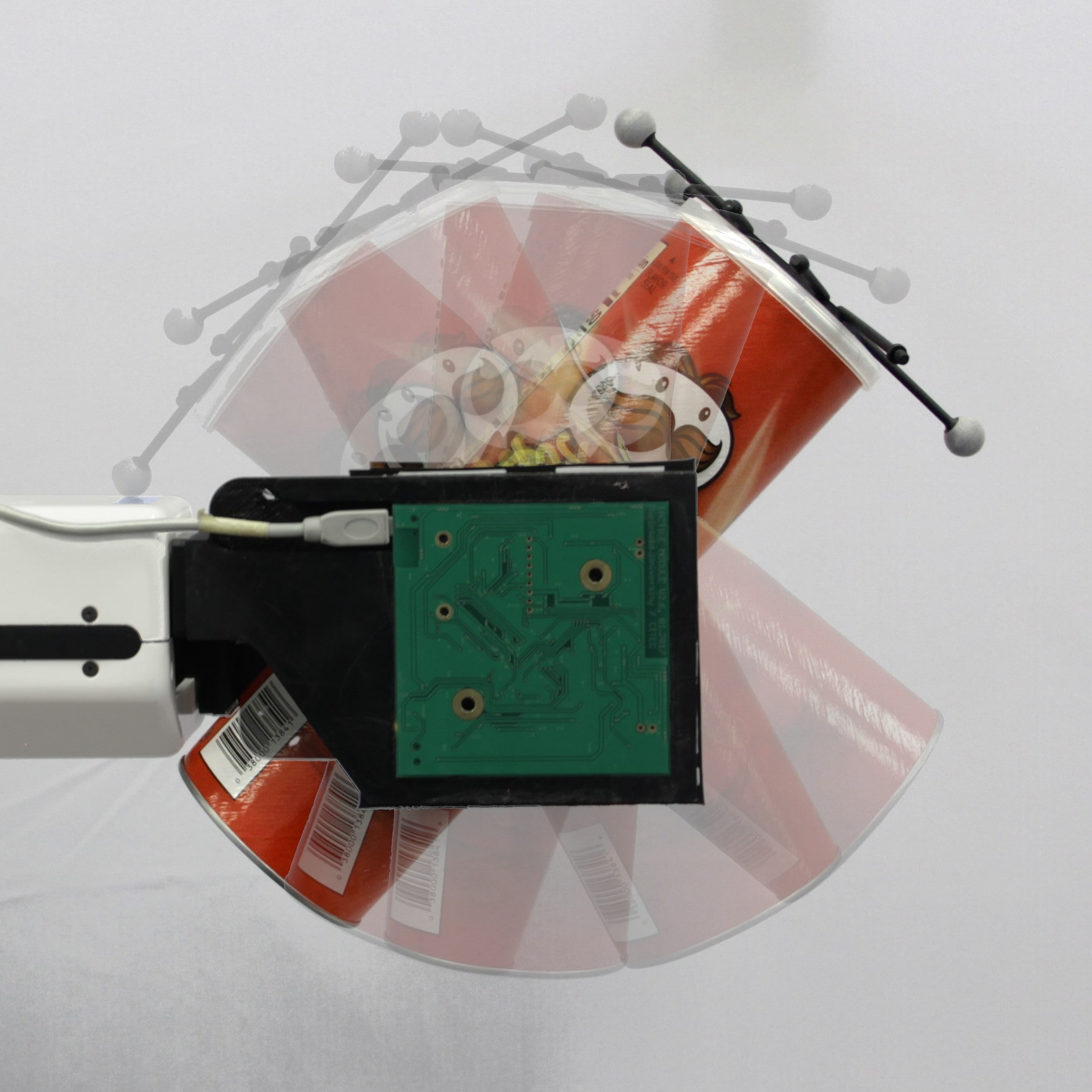}}
\hfill
\vspace{-0.27cm}
\captionof{figure}{Experiment objects and object poses used for evaluation and data collection.}
\vspace{-0.15cm}
\begin{center}
\captionof{table}{\small Training results of networks with different input sensor modalities. We report the lowest test loss averaged over 10 batches.} 
\label{table:test_loss}
\scriptsize
\begin{tabular}{l|ccc}
     & Tactile & F/T &  Tactile + F/T  \\
\hline
\hline
Test Loss (rad)   & \textbf{0.03} & 0.43 & 0.05 \\
\hline
\end{tabular}
\end{center}
\vspace{-0.5cm}
\end{figure}

We used two primitive 3D-printed objects for data collection, i.e., 
a cylinder ($2.25$cm radius, $15$cm length) and a cuboid ($5\times5\times19$cm) as shown on the left in Fig. \ref{fig:objects}.
To collect the ground-truth orientation that is required to train our networks, we use OptiTrack, an external, infra-red camera-based marker tracking system.
Thus, during data collection, markers were attached to both, robot and object.

For training-data generation, the objects were randomly positioned in the gripper and then grasped using the force controller from \cite{lach2022force}.
The robot's arm was then manually brought into a realistic initial pose using gravity compensation mode.
Then, a sample was collected by storing tactile data, F/T data, $\Rgop$, $\Rwg$ and $\zgt$.
For each arm pose, the object's in-hand pose was varied $10$ times within a range of $160^\circ$ corresponding to plausible grasp poses.
Afterward, the arm's pose was changed, and the process was repeated.
Using this method, we collected 800 samples per object and $1600$ samples in total. We trained every neural net architecture for $40$ epochs and reserved $20\%$ of the data for testing.
The network parameters were stored every time the running average of the test loss over the last $10$ batches was lower than before.

Table \ref{table:test_loss} shows the lowest average test loss for all three architectures.
The tactile-only model achieved the best result, although only marginally outperforming the tactile + F/T model with its test performance being only lower by $~1.1^\circ$.
The F/T-only model performed significantly worse with an average angular error of $0.43$ rad, indicating that it might not have learned the task sufficiently well.

\subsection{Line-fitting baselines}
To assess whether a data-driven model is required to solve this task robustly and to gauge its performance compared to other approaches, we introduce two baseline models for comparison.
Given the nature of our sensory data and the goal to find the object's main axis within it, line-fitting methods naturally come to mind.
We chose two popular methods from that field, Principal Component Analysis (PCA) \cite{pca} and Hough transforms \cite{duda1972use}.
As both methods work on individual images, we combine the two sensor readings into one frame by flipping one of the sensor images to account for symmetry.
Therefore, the input to the baselines contains the information of both sensors. This should increase robustness as the sensors might be differently affected by noise.

\subsubsection{PCA baseline}
We treat the force readings as a bi-variate, uni-modal Gaussian and estimate its mean, standard deviations, and covariance matrix $\mathbf{C}$.
To obtain the orientation of the object's main axis, we calculate the first principal component of $\mathbf{C}$ using PCA.
Assuming that an object's main axis lies along its largest grasping surface, the first principal component should constitute a decent estimation for said axis.
We then calculate the angle of the line relative to the sensor and transform it into the rotation $\Rgop$, allowing us to generate a corrective motion that aligns the objects.

\subsubsection{Hough transform baseline}
Hough transform is a common tool in image processing for finding lines in images. 
Typically, a raw input image undergoes some preprocessing where edges are extracted, and finally, the Hough transform is applied to the resulting binary image.
Empirically, we have found the Hough transform to show better performance if we simply create a binary image by assigning a value of $1$ to taxels with a force above a noise threshold and $0$ otherwise.
We only consider the resulting line with the most votes (the most confident estimate).
Lines are parametrized by the angle $\psi$ between the $x$-axis and a line normal that intersects the origin and the distance to the origin of said normal.
From $\psi$, we calculate the angle between the $x$-axis and the line itself $\phi = \pi - \psi$ and again calculate $\Rgop$ as with the other baseline. 

\subsection{Experimental setup}

For the real-world experimental evaluation of placing success, we let each method place an object from 5 different arm poses and 4 different in-hand object poses, yielding 20 samples per object for each method.
Fig. \ref{fig:evalposes} shows the Pringles object in four exemplary in-hand object poses that were used during evaluation.
During the evaluation, we again attach markers to the objects to assess each method's performance w.r.t. ground truth.
Additionally, we execute the placing controller (cf. Sec. III-C) for each trial and note whether the object was placed successfully, i.e. did not fall or tip over after opening the gripper.
Situations, where the object tilts but is still supported by the gripper, are also considered a failure.

\subsection{Ablation over sensor modalities with seen objects}

\begin{table}[t]
\begin{center}
\caption{\small Comparison of different placing methods on the two 3D-printed training objects (cf. Fig. \ref{fig:objects}). The angular error is the angle between the predicted $\zp$ and the one measured with OptiTrack.} 
\label{table:train_eval}
\scriptsize
\scalebox{1.0}{
\begin{tabular}{l|c||cc||c}
Method & Metric  & Cylinder &  Cuboid & Average \\
\hline 
\hline

\multirow{2}{*}{OptiTrack} & 
    \multirow{2}{*}{\shortstack{ \% Suc. \\ Ang.Err. }} &
    \multirow{2}{*}{\shortstack{ 90\%   \\ - }} & 
    \multirow{2}{*}{\shortstack{ \textbf{95\%}  \\ - }} &
    \multirow{2}{*}{\shortstack{ \textbf{92.5}\% \\ - }}\\
&&&&\\
\hline

\multirow{2}{*}{Tactile} & 
    \multirow{2}{*}{\shortstack{ \% Suc. \\ Ang.Err. }} &
    \multirow{2}{*}{\shortstack{ \textbf{95\%} \\ \textbf{0.06 $\pm$ 0.03}  }} & 
    \multirow{2}{*}{\shortstack{ 85\% \\ 0.17 $\pm$ 0.12 }} &
    \multirow{2}{*}{\shortstack{ 90.0\% \\ \textbf{0.11 $\pm$ 0.08} }}\\
&&&&\\
\hline

\multirow{2}{*}{Tactile + F/T} & 
    \multirow{2}{*}{\shortstack{ \% Suc. \\ Ang.Err. }} &
    \multirow{2}{*}{\shortstack{ 90\% \\ 0.08 $\pm$ 0.04  }} & 
    \multirow{2}{*}{\shortstack{ 75\% \\ \textbf{0.16 $\pm$ 0.19} }} &
    \multirow{2}{*}{\shortstack{ 87.5\% \\ 0.12 $\pm$ 0.08 }}\\
&&&&\\
\hline

\multirow{2}{*}{F/T} & 
    \multirow{2}{*}{\shortstack{ \% Suc. \\ Ang.Err. }} &
    \multirow{2}{*}{\shortstack{ 15\% \\ 0.38 $\pm$ 0.26 }} & 
    \multirow{2}{*}{\shortstack{ 25\% \\ 0.39 $\pm$ 0.32 }} &
    \multirow{2}{*}{\shortstack{ 20.0\% \\ 0.38 $\pm$ 0.29 }}\\
&&&&\\
\hline

\multirow{2}{*}{PCA} & 
    \multirow{2}{*}{\shortstack{ \% Suc. \\ Ang.Err. }} &
    \multirow{2}{*}{\shortstack{ 90\% \\ 0.07 $\pm$ 0.02 }} & 
    \multirow{2}{*}{\shortstack{ 10\%  \\ 0.83 $\pm$ 0.42 }} &
    \multirow{2}{*}{\shortstack{ 50\% \\ 0.45 $\pm$ 0.22 }}\\
&&&\\
\hline

\multirow{2}{*}{Hough} & 
    \multirow{2}{*}{\shortstack{ \% Suc. \\ Ang.Err. }} &
    \multirow{2}{*}{\shortstack{ 80\% \\ 0.09 $\pm$ 0.04 }} & 
    \multirow{2}{*}{\shortstack{ 10\% \\ 0.76 $\pm$ 0.37 }} &
    \multirow{2}{*}{\shortstack{ 45\% \\ 0.42 $\pm$ 0.21 }}\\
&&&\\
\hline

\end{tabular}}
\end{center}
\vspace{-0.7cm}
\end{table}

We first compare the OptiTrack baseline with all three networks and the two line-fitting baselines on the objects used for training.
Thus, we evaluate $6$ methods on the two 3D-printed training objects and conduct $20$ trials per object, resulting in $240$ placing trials in total.
Table \ref{table:train_eval} shows the results. 
Note, that we consider the OptiTrack measurements as the baseline since it was not possible to obtain more precise object state estimations.
We can compare the OptiTrack baseline to the other methods based on success rate and the angular error between their predictions of the object normal and OptiTrack's ground-truth measurement for $\zpg$ (c.f.~eq.~\ref{eq:loss}).

A surprising result is that the tactile-only model performed better in terms of a success rate than the OptiTrack baseline.
This can be explained by OptiTrack not measuring the marker positions perfectly and accurately all the time.
When the robot is blocking the sight of some cameras on the markers, this leads to a degradation in measurement accuracy or even an interruption of tracking. 
Additionally, the cylinder was quite top-heavy, which requires a relatively precise alignment with the table for it to not topple.
For the cuboid object, which is not as challenging to place safely, the OptiTrack baseline performed better than the tactile model, with 95\% vs. 85\% success rate, respectively.

Among the neural networks, the tactile-only model performed best in almost all metrics.
Only in terms of estimation accuracy for the cuboid object did the tactile+F/T model achieve higher scores, however only negligibly so by $0.01$ rad.
It also exhibited a higher variance in accuracy as compared to the purely tactile model, resulting in less reliable results in terms of success rate.
The evaluation also confirmed another indication from training, namely that the F/T-only model did not succeed in estimating the object's placing normal sufficiently well for stable placing.
In most cases, it simply predicted the identity matrix resulting in no corrective movements.
Following a similar argumentation as in \cite{li2020review}, we posit that the object's mass might be negligible compared to the actual weight of the end-effector. Thus, F/T signals alone do not seem to carry enough information for assessing the relative transform between gripper and the object's placing normal.

PCA has comparable performance to the tactile-based neural networks for the cylindrical object and showed the lowest variation there, with Hough transforms performing slightly worse, yet satisfactory.
For the cuboid object, however, the performance decreased sharply for both classical methods.
While cylindrical objects generate an easily identifiable, unique line in the sensor images, cuboid ones cover more contact areas and create less distinct patterns.
The neural networks still appear to identify the relevant patterns robustly, while PCA and Hough heavily suffer from this fact.
Different from the F/T-only network, the classical approaches do not involve a training phase and, thus, they cannot learn some structure of the data and generalize over them, as effectively done by our neural network approach.
While the network internalized a range of plausible angle offsets, while learning to filter-out noise, PCA and Hough often produce severely wrong estimates, despite our preprocessing efforts, resulting in large errors and only a few successful placing trials.

\begin{table*}[t]
\vspace{0.2cm}
\begin{center}
\caption{\small Experimental results for five household testing objects (cf. Fig. \ref{fig:objects}). All objects were unknown to the models before this evaluation, i.e., they were not present in the training set. Angular errors are reported in radians.} 
\label{table:test_eval}
\scriptsize
\scalebox{1.0}{
\begin{tabular}{l|c||c|c|c|c|c|c|c||c}

Model & Metric & Pringles &  Glue Bottle & Tabasco & Mallow Pop & Cheez It & Shampoo & Lipstick & Average\\
\hline
\hline

\multirow{2}{*}{Tactile} & 
    \multirow{2}{*}{\shortstack{ \% Suc. \\ Ang.Err. }} &
    \multirow{2}{*}{\shortstack{ \textbf{90}\%  \\ \textbf{0.07 $\pm$ 0.03} }} & 
    \multirow{2}{*}{\shortstack{ 85\%           \\ 0.08 $\pm$ 0.04 }} & 
    \multirow{2}{*}{\shortstack{ \textbf{85}\%  \\ \textbf{0.10 $\pm$ 0.13} }} & 
    \multirow{2}{*}{\shortstack{ \textbf{80}\%  \\ \textbf{0.16 $\pm$ 0.10} }} & 
    \multirow{2}{*}{\shortstack{ 80\%           \\ 0.10 $\pm$ 0.06 }} & 
    \multirow{2}{*}{\shortstack{  \textbf{85}\% \\ \textbf{0.06 $\pm$ 0.03} }} &
    \multirow{2}{*}{\shortstack{ \textbf{90}\% \\ -}} &
    \multirow{2}{*}{\shortstack{ \textbf{85}\% \\ \textbf{0.09 $\pm$ 0.07} }}\\
&&&&&&&&\\
\hline

\multirow{2}{*}{Tactile + F/T} & 
    \multirow{2}{*}{\shortstack{ \% Suc. \\ Ang.Err. }} &
    \multirow{2}{*}{\shortstack{ 85\%   \\ 0.13 $\pm$ 0.20 }} & 
    \multirow{2}{*}{\shortstack{ 80\%  \\ 0.09 $\pm$ 0.04 }} & 
    \multirow{2}{*}{\shortstack{ 70\% \\ 0.16 $\pm$ 0.10 }} & 
    \multirow{2}{*}{\shortstack{ 70\% \\ 0.14 $\pm$ 0.10 }} & 
    \multirow{2}{*}{\shortstack{ \textbf{85}\% \\ \textbf{0.05 $\pm$ 0.06} }} & 
    \multirow{2}{*}{\shortstack{ 75\% \\ 0.10 $\pm$ 0.06 }} & 
    \multirow{2}{*}{\shortstack{ 80\% \\ - }} &
    \multirow{2}{*}{\shortstack{ 77\% \\ 0.12 $\pm$ 0.10}}\\
&&&&&&&&\\
\hline

\multirow{2}{*}{PCA} & 
    \multirow{2}{*}{\shortstack{ \% Suc. \\ Ang.Err. }} &
    \multirow{2}{*}{\shortstack{  \textbf{90}\% \\ 0.08 $\pm$ 0.06  }} & 
    \multirow{2}{*}{\shortstack{  \textbf{90}\% \\ \textbf{0.07 $\pm$ 0.03}  }} & 
    \multirow{2}{*}{\shortstack{  15\%          \\ 0.79 $\pm$ 0.61  }} & 
    \multirow{2}{*}{\shortstack{  20\%          \\ 0.70 $\pm$ 0.45  }} & 
    \multirow{2}{*}{\shortstack{  80\%          \\ 0.09 $\pm$ 0.05  }} & 
    \multirow{2}{*}{\shortstack{  75\%          \\ 0.08 $\pm$ 0.0  }} & 
    \multirow{2}{*}{\shortstack{  85\% \\ -  }} &
    \multirow{2}{*}{\shortstack{  65\% \\ 0.30 $\pm$ 0.24 }}\\
&&&&&&&&\\
\hline

\multirow{2}{*}{Hough} & 
    \multirow{2}{*}{\shortstack{ \% Suc. \\ Ang.Err. }} &
    \multirow{2}{*}{\shortstack{  85\%          \\ 0.11 $\pm$ 0.06  }} & 
    \multirow{2}{*}{\shortstack{  80\%          \\ 0.10 $\pm$ 0.03 }} & 
    \multirow{2}{*}{\shortstack{  60\%          \\ 0.20 $\pm$ 0.30  }} & 
    \multirow{2}{*}{\shortstack{  50\%          \\ 0.35 $\pm$ 0.37   }} & 
    \multirow{2}{*}{\shortstack{  70\%          \\ 0.12 $\pm$ 0.08   }} & 
    \multirow{2}{*}{\shortstack{  70\%          \\ 0.16 $\pm$ 0.32  }} & 
    \multirow{2}{*}{\shortstack{  80\% \\ -  }} &
    \multirow{2}{*}{\shortstack{  70\% \\ 0.17 $\pm$ 0.17 }}\\
&&&&&&&&\\
\hline 
\end{tabular}}
\end{center}
\vspace{-0.75cm}
\end{table*}

\subsection{Evaluation with previously unseen objects}
\begin{figure}[t!]
    \centering
    \includegraphics[width=0.44\textwidth]{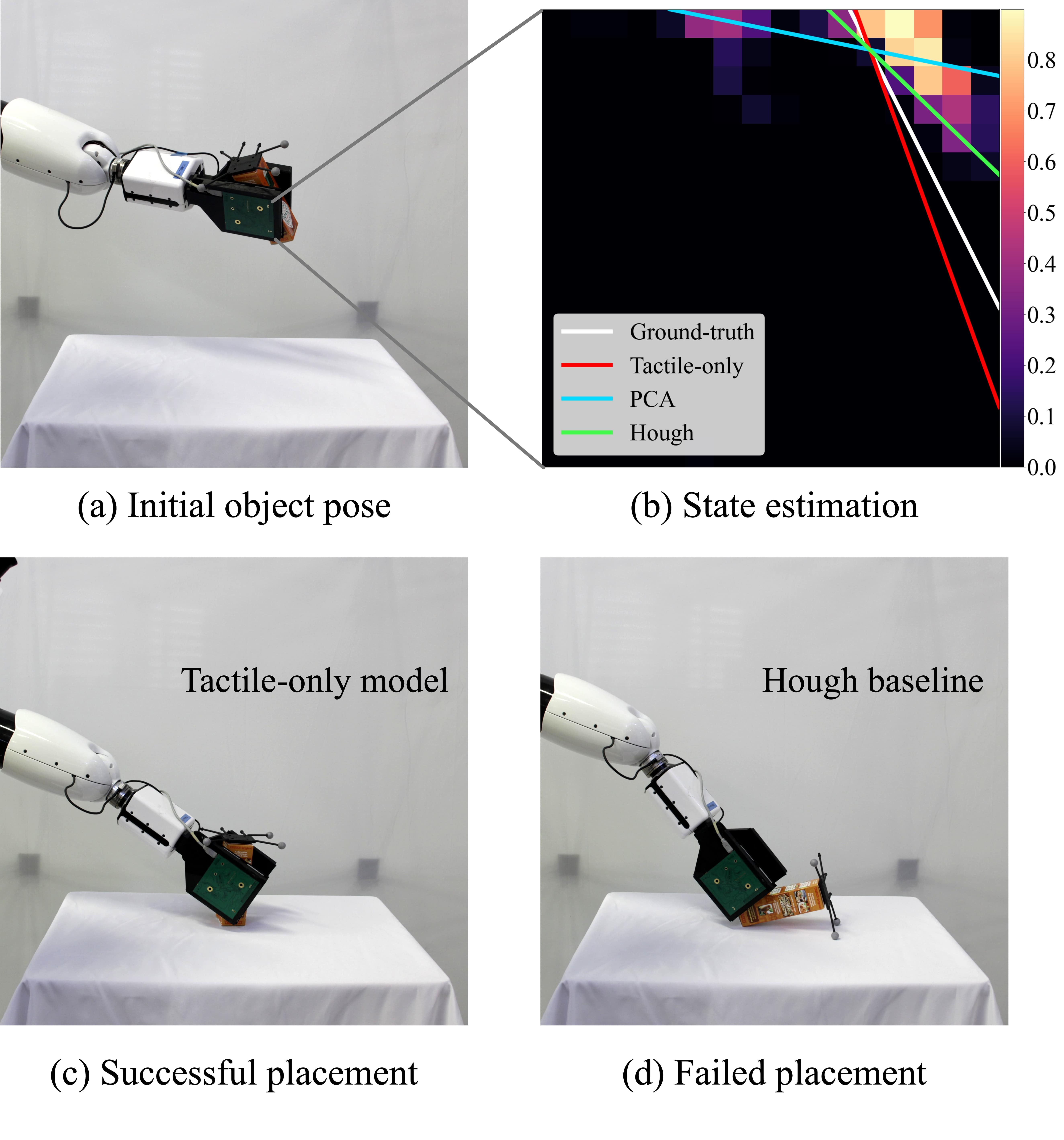}
    \caption{\small Placement sequence comparing the tactile-only model to the two baselines. The network performed best with an angular error of 0.11 and placed the Tabasco correctly, while Hough and PCA both failed with estimation errors of 0.34 and 0.91 respectively. (b) depicts the force intensities for each taxel along with each model's prediction of the object's placing normal and the ground 
    -truth obtained from OptiTrack.}
    \label{fig:placing_seq}
    \vspace{-0.65cm}
\end{figure}
After confirming the training results on the real robot in our first experiment, we conducted a second evaluation on objects that were not present in the training data.
During this evaluation, we also attached markers to the objects to validate the prediction accuracy.
We do not report prediction accuracy for the lipstick as attaching a marker ensemble would substantially alter its placing dynamics and, thus, report solely the success rate.
We only evaluated the two best neural models from the previous experiment, namely the tactile-only and the tactile with F/T neural nets, along with the two classical approaches.
We evaluated the $4$ methods on $7$ different household objects for 20 trials each, hence performing 560 placing trials in total.
Two of the objects were cylindrical (Glue Bottle \& Pringles), three objects were box-like (Mallow Pop, Tabasco \& Cheez-It), and the Lipstick was a small, elongated, rectangular object with rounded edges (see Fig. \ref{fig:objects} on the right side).

The results of this evaluation are given in Table \ref{table:test_eval}. 
We, again, report the success rate of correct placements and the angular error between the model's predictions and OptiTrack where possible.
The network models perform very well on most unknown objects, indicating that our method generalizes across object primitives of unknown dimensions.
Similar to the results from the previous evaluation on known objects, the tactile-only model showed superior performance in most cases, as it performed best on average in both metrics with a low variance in results.
The tactile {+} F/T model only performed better in terms of both metrics when placing the Cheez-It box.
We hypothesize that the slightly worse performance might result from the network receiving rather non-informative F/T signals alongside the tactile data.

The PCA baseline showed similar performance to the neural networks on cylindrical objects.
As in the first experiment, its performance dropped sharply on box-like objects, which can likely be attributed to a less pronounced distribution of forces around the object's main axis.
Lastly, the Hough model performed well to mediocre on all evaluation objects.
While it did not exhibit such a drastic performance drop on box-like objects as PCA, it did not manage to achieve satisfactory results consistently.
Nevertheless, it performed better on box-like objects than on the cuboid test object. 
Paper boxes tend to create more distinct lines (one at each each of the object) which are easier to detect with Hough transforms, whereas the much more rigid 3D-printed cuboid creates a larger contact patch, making it more difficult to detect lines.
Fig.~\ref{fig:placing_seq} shows placing trials of the box-like Tabasco object, where the tactile-only model successfully placed the object whereas both classical models' predictions led to failed placements.
Finally, all models showed reasonable to good performance on the Lipstick, an object that is difficult to place due to its small placing faces.
The tactile-only network performed best among all methods, further underlining its generalizability.

\subsection{Discussion}

Our results show that tactile data is a crucial source of information for predicting the object's placing normals, whereas F/T data has not been proven to be as informative for this task.
Furthermore, our evaluation has shown that neural network models generalize to unknown objects with high success rates and precision. 
It also revealed that classical approaches can work reliably on a set of objects with specific characteristics (cylinders) while not being accurate enough on objects that lack these (boxes).
This leads us to the conclusion that, despite the effort of data collection, neural networks are the most sensible model choice for this task due to their high accuracy and robustness in real robotic experiments with unseen objects, of different sizes and weights than the ones used during training.

On the other hand, our evaluation also revealed some limitations of our approach.
Our method cannot deal well with objects that uniformly cover the complete sensor surface.
To circumvent this problem, the ability of in-hand object manipulation would be required to again recover more distinctive tactile feedback. 
Yet, such maneuvers are severely limited by our parallel gripper. 
Thus, we hypothesize that future work should aim at executing a larger variety of exploratory object interactions to acquire better tactile signals and enable robust object placing with an even wider range of objects.
Another limitation arises from our assumption of downward-pointing placing faces.
In situations where this assumption is likely to be violated, tactile data could be used to fine-tune higher-level object state predictions, e.g., from a vision system with less accuracy.

%% file: sections_new/conclusions.tex
\section{Conclusions}

In this work, we investigated the potential of employing tactile sensing for obtaining stable object placing policies.
This task is especially challenging as we neither assume any prior knowledge about the object's initial pose, nor any access to external camera-based sensing.
We presented an approach based on deep learning that takes as input tactile or F/T data, or a combination of both, and predicts a rotation matrix that is supposed to align the object's placing normal with that of the placing surface. 
We trained a representation through supervised learning with a small dataset of primitive objects.
In the first part of the experimental evaluation, we validated our approach using the training objects and concluded that F/T sensing alone is not sufficient for this task.
Our subsequent evaluation of previously unseen household objects showed an impressive generalization of our approach.
We also compared the neural network models to classical, learning-free methods, and found that our proposed tactile-based neural policy shows higher performance in general and better generalization properties, justifying the choice of a learning-based method.
In the future, we intend to investigate a more interactive method for stable object placing through active touch.